\def\year{2018}\relax
\documentclass[letterpaper]{article} %
\usepackage{aaai18}  %
\usepackage{times}  %
\usepackage{helvet}  %
\usepackage{courier}  %
\usepackage{url}  %
\usepackage{graphicx}  %
\begin{document}
\title{Transfer Learning in CNNs Using Filter-Trees}
\author{Suresh Kirthi Kumaraswamy, PS Sastry \and KR Ramakrishnan\\
{\tt\small kirthifame@gmail.com},
{\tt\small sastry@iisc.ac.in},
{\tt\small krr2504@gmail.com} \\
Indian Institute of Science\\
Bangalore, India\\
}
\maketitle
\begin{abstract}
Convolutional Neural Networks (CNNs) are very effective for many pattern recognition tasks. However, training  deep CNNs needs extensive computation and large training data. In this paper we propose \textit{Bank of Filter-Trees}~(BFT) as a transfer learning mechanism for improving efficiency of learning CNNs. A filter-tree corresponding to a filter in $k^{th}$ convolutional layer of a CNN is a subnetwork consisting of the filter along with all its connections to filters in all preceding layers. An ensemble of such filter-trees created from the $k^{th}$ layers of many CNNs learnt on different but related tasks, forms the BFT. To learn a new CNN, we sample from the BFT to select a set of filter trees. This fixes the target net up to the $k^{th}$ layer and only the remaining network would be learnt using training data of new task. Through simulations we demonstrate the effectiveness of this idea of BFT. This method constitutes a novel transfer learning technique where transfer is at a subnetwork level; transfer can be effected from multiple source networks; and, with no finetuning of the transferred weights, the performance achieved is on par with networks that are trained from scratch. 
\end{abstract}

\section{Introduction}
\label{sec:intro}

\noindent 
Deep learning has been highly successful in tackling many machine learning problems in 
vision, speech, text and many other areas. The Convolutional Neural Networks~(CNNs) represent a deep neural network architecture that is the de facto standard for many image recognition tasks today. While CNNs are very effective in many applications, training of CNNs requires a large number of labelled examples as well as a large amount of computing resources. 
Transfer Learning is a popular approach suitable for improving the efficiency of learning~\cite{silver2008inductive,sharkey1993adaptive}.  
A simple transfer learning protocol for CNNs would be as follows. We start with a CNN that was previously trained
on a related task and then finetune the weights using the training examples for the new target task. Finetuning of weights involves learning weights using the same learning algorithm. However, this way of initializing the weights of the target CNN using the weights of a previously learnt CNN (rather than initializing the weights randomly) results in more efficient learning both in terms of number of training epochs and the number of training examples needed.

There are many ways in which this basic idea of transfer learning can be improved. The transfer learning as outlined above essentially amounts to only a good heuristic for initializing the weights. It is more interesting to see if some aspects of what is learnt earlier can be directly used in the new target task. We may have learnt many CNNs previously on different tasks, all of which are somewhat related to the new target task.  Hence it would be nice to have a generic mechanism whereby one can transfer knowledge (weights) from multiple learnt CNNs to a new target net. In the normal transfer learning method, the architecture of the target net has to be same as (or very similar to) that of the previously learnt CNN. Transfer learning can be much more effective if this constraint can be removed. One of the main motivations for deep networks, in general, and CNNs in particular, is that we automatically learn the relevant features from the data. The filters in the convolutional layers of a CNN essentially represent such learnt features. It is reasonable to expect that there would be a generic set of features that should be useful in a host of visual pattern recognition tasks. Given a set of already learnt CNNs, the learnt features should be useful in  new  tasks also. Thus, ideally, transfer learning should be a mechanism for reusing previously learnt features for a new task (rather than being a heuristic for initializing weights).

In this paper we propose a mechanism that we call \textit{bank of filter-trees}~(BFT) which can potentially address all the issues outlined above. The term \textit{filter-tree}, refers to a tree or a subnetwork that we associate with a filter in any convolutional layer of a CNN. The filter-tree of a filter in \textit{layer}-$k$ of a CNN contains the filter itself along with all the filters that effectively connect to it from all the preceding $(k-1)$ layers. Given any (learnt) CNN, we can create a set of such filter-trees corresponding to individual filters in any convolutional layer. If we have a number of pre-learnt CNNs, we can do the same for each CNN and this ensemble of all filter-trees is what we call \textit{bank of filter-trees}~(BFT). In our method of transfer learning, the elements of this BFT are what are considered the exchangeable elements. Suppose we have a BFT containing each of the \textit{layer-k} filters of all the learnt CNNs. We can now create a new CNN where \textit{layer-k} (and hence all the earlier layers) are populated by some selected (or randomly sampled) elements from the BFT and all higher layers would have randomly initialized weights which need to be learnt. (The weights in the subnetwork populated by elements from the BFT are fixed and they are not finetuned for the target task). This is a mechanism of transfer learning where individual features that we learnt are what are being transferred. The transfer can be effected from several pre-learnt CNNs in a seamless manner. The architecture of the new or target CNN can be different from earlier learnt CNNs. Thus, this idea has the potential to address all issues we outlined earlier. 
In this paper we formulate the idea of BFT and explain how it can be implemented. We also present empirical investigations which clearly demonstrate the effectiveness and efficiency of this type of transfer learning. To the best of our knowledge, this is the first instance of a transfer learning method where transfer is at a subnetwork level; transfer can be effected from multiple source networks; and, with no finetuning of the transferred weights, the performance is on par with that of normally learnt networks.

\begin{figure*}
\begin{center}
  \includegraphics[scale=0.175]{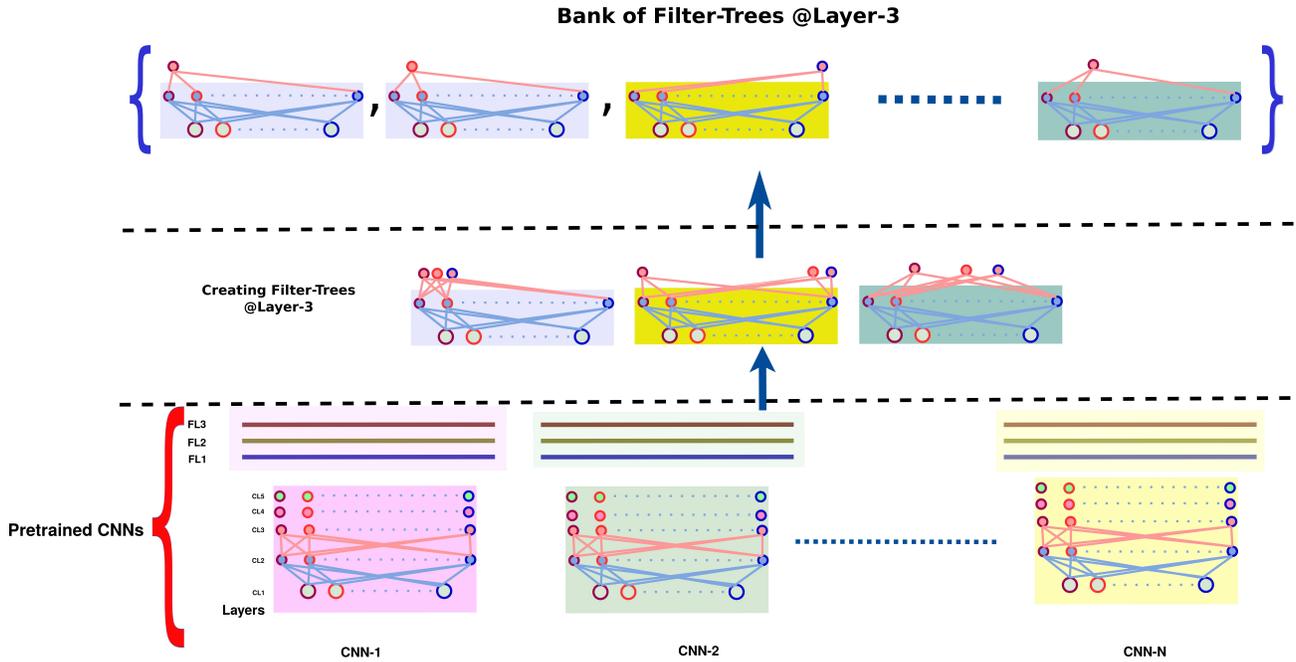}
\center{\caption{Illustrating the building of a BFT at the level of convolutional layer-3 using \textbf{\textit{N}}-\textit{number} of CNNs each with \textit{five-convolutional} and 
\textit{three-fully-connected} layers. From these \textit{N-pretrained CNNs} we tweeze out 
filters in the layer-3. From each CNN  and each filter, we build a tree of filters with the filter at the layer-3 on the top and containing all the filters at the \textit{layer-1 and 2} along with  
 their connections. These filter-trees form the BFT at layer-3. \textit{(Note:This figure is best viewed in color)} \label{fig:fig_BFT}}}
\end{center}

\end{figure*}

\subsection{Related Work}
Transfer learning can be viewed as any method that uses pre-learnt networks for improving efficiency of learning a new task. A straight-forward method is to start with a pre-learnt network (that is, take the learnt weights as the initial weights) and finetune all weights using the training data of the target task, resulting in more efficient learning~\cite{zhou2014learning}. One can also allow for some variation in the architecture. For example, one can start with a learnt network and either add one or more layers on top of it or expand a layer by adding a few randomly initialized nodes to it~\cite{oquab2014learning,wei2016network}. Unlike these, where we start with a single pretrained network, in the method proposed here an ensemble of filter-trees is created by pooling together any number of pre-trained networks. The architecture of the target network can be very different from that of any of the source networks. Also, in our method, the weights transferred are fixed and are not finetuned. 

There are transfer learning methods reported in literature where previously learnt features are directly used. In~\cite{sharif}, the convolutional layers of a learnt CNN are used as fixed feature extractors for the new task and the training data of the new task, represented as feature vectors like this, is then used for training a new classifier, e.g., an SVM. In our method, the filter-trees represent pre-learnt features; but they come from multiple source networks. Also, since the target network can have one or more convolutional layers on top of the filter-trees, in the new task we can learn useful ways of combining previously learnt features.  Some earlier works in transfer learning in neural networks such as 
~\cite{pratt1991direct,pratt1993discriminability} also use multiple source networks for transfer. However, here, each of these networks is previously trained on a subtask of the target task and then these networks are simply fused together to learn the whole of the target task. Our method based on an ensemble of filter-trees is a much more general method of effecting transfer learning from multiple source networks. 

Transfer learning has also been used as method to augment the data set of the target task. If the target task has less number of training examples, one can use pre-learnt CNNs as feature extractors to decide which of the patterns from some other task are similar to the training data of the target task and thus increase the number of training examples. (See, for example, \cite{ge2017borrowing}). Though the previously learnt feature generators are used here to aid in the learning of a new task, it is quite different from the type of transfer learning that is explored in this paper. Another similar transfer learning method, that relies on determining which sets of examples are similar, is reported in~\cite{srivastava2013discriminative}. This paper also talks about a tree structure for transfer learning though it is quite different from our use of filter-tree here. The tree in~\cite{srivastava2013discriminative} refers to tree of object categories which is an externally supplied information which allows one to decide on the distances between different sub-categories of examples. 

The question of the level or granularity at which learnt weights can be transferred in CNNs has also been investigated. Transfer learning where an entire convolutional layer can be transferred from a source network to a target network is explored in~\cite{yosinski}. The method presented in~\cite{kirthi} is similar to our method and it also uses a bank of filter weights from multiple source networks to effect transfer and they explore transfer learning both at convolutional layer level as well as at the level of individual filters of learnt networks. However, these methods need finetuning of all weights to reach the performance achieved by networks trained from scratch. This is because individual weight vectors of filters in the pre-learnt networks, by themselves, do not contain the full context of the features learnt earlier. The filter-trees that we propose here seem to have the correct granularity for transferring the earlier learnt knowledge of useful features as demonstrated by the results we present here. 

Our method involves sampling from an ensemble of pre-learnt filters to populate the convolutional layers of a target network and hence it may seem similar to classifier ensemble methods such as bagging or boosting~\cite{breiman1996bagging,krogh1995neural,schapire2012boosting}. However in such methods all networks in the ensemble are trained on the same task whereas in our method, the ensemble of filter trees are created from networks trained on different tasks and the transfer learning is for a new task.   

This work presented here can be viewed as a step in the direction of creating an {\em universal bank of feature generators} which can be used as a dictionary to realize CNNs across tasks. 
Several studies in neuroscience have pointed out that representations that brain seems to be using for visual or text stimuli are highly similar across individuals~\cite{mitchell,dicarlo,neuron}. This lends support to the existence of some universal set of feature generators that should be useful for classification of sensory signals. The bank of filter-trees proposed here explores this idea in the limited context of transfer learning in CNNs. 

The rest of the paper is organized as follows. Section~\ref{sec:bft} introduces the concept of Bank of filter-trees.
Section~\ref{sec:expt-setup} describes our experimental setup. Section~\ref{sec:results} discusses the experimental results showing the performance of the BFT based transfer learning and discusses why filter-trees as an unit of transfer (as opposed to individual filters as a unit of transfer) is better. We conclude the paper in Section~\ref{sec:conc}.

\section{Bank of Filter-Trees}
\label{sec:bft}

We associate what we call a filter-tree with any filter in any convolutional layer of a CNN. The filter-tree of a filter in the $k^{th}$ layer is a subnetwork consisting of the filter, all its connections to filters in layer-$(k-1)$, and the connections of these filters to filters in its previous layer and so on till the input layer. Thus a filter-tree of a filter essentially represents a complete (learnt) feature extractor (corresponding to that filter).

Consider a CNN with $L$ convolutional layers and with $N^l$ filter in layer-$l$, $l=1, \cdots, L$. Let $f_{k}^{l}$ denote the $k^{th}$ filter in layer-$l$, $k=1, \cdots, N^l$. Let $f^j_{1:N^j}$ represent the set of all filters in layer-$j$. Then the filter-tree, $\mathbf{T_{{f}_{k}^l}}$, corresponding to filter $f_{k}^{l}$, 
 can be symbolically represented as $\lbrace {f}_{1:N^1}^1$ $\rightarrow$ ${f}_{1:N^2}^2$ 
$\rightarrow$ .....$\rightarrow$ ${f}_{k}^l \rbrace$.

We can fix some $k$ and extract such filter-trees corresponding to all filters in layer-$k$ of a pre-learnt CNN. When we have learnt many CNNs and want to use all of them as source networks, we get these filter-trees from each of the learnt CNNs. This ensemble of filter-trees constitutes what we term as bank of filter-trees~(BFT). 

Figure~\ref{fig:fig_BFT} schematically illustrates this process of constructing a BFT.

\subsection{Initialization Using BFT}
To construct a new CNN for a given target task, we need to sample sufficient number of filter-trees from the BFT. Ideally we should select the filters so that we maximize diversity and minimize redundancy among the selected set of filters. For the results reported in this paper, we just randomly sample (without replacement) from the BFT. Thus, our process of constructing a new CNN using the BFT is as follows. 

Suppose our BFT consists of filter-trees at layer-$k$. For the target network architecture, we decide first on the number of filters in layer-$k$. Then we randomly choose that many filter-trees from BFT. This fixes the target network up to layer-$k$. (please also see the discussion below). We are completely free to choose the architecture of the target network above layer-$k$ and the weights in all these layers are initialized randomly. 

As is easy to see, when we populate the layer-$k$ of the target CNN with filter-trees, it automatically populates the first $(k-1)$ layers also. If two filter-trees sampled from the BFT happen to be from the same source CNN, then they share the filters in the first $(k-1)$ layers and hence the two filter-trees can be fused together to form a proper subnetwork. This would be done for all filter-trees coming from any one source network. If we had $M$ source networks from which the BFT is formed, then the number of filters in the earlier layers of the target network would be sum of the number of filters of the source networks. However, this does not really increase the computational burden. Note that in the target network constructed using BFT, all the weights up to this layer-$k$ are fixed and they are not learnt. While the memory for storing the weights increases, this increase is only additive and not multiplicative. Suppose our BFT is constructed using layer-3 filters from $M$ source networks each having $N$ filters in layer-2. Suppose we chose to use $n$ filters in the layer-3 of target network. Now the layer-2 of target network would have at most $MN$ filters. However, each of the $n$ filters in the layer-3 of target network can connect to only $N$ filters because these connections come from one of the source networks. Our experimental results show that a BFT constructed from just four source networks is also quite effective for efficient transfer learning. Thus, in practice, the extra memory needed is unlikely to be problematic. 

Finally the target network is trained using the examples of the new task. In this training only the weights in layer-$(k+1)$ onwards are learnt; the weights in the first $k$ layers are fixed as explained above.

\begin{figure*}
\begin{center}
  \includegraphics[scale=0.21]{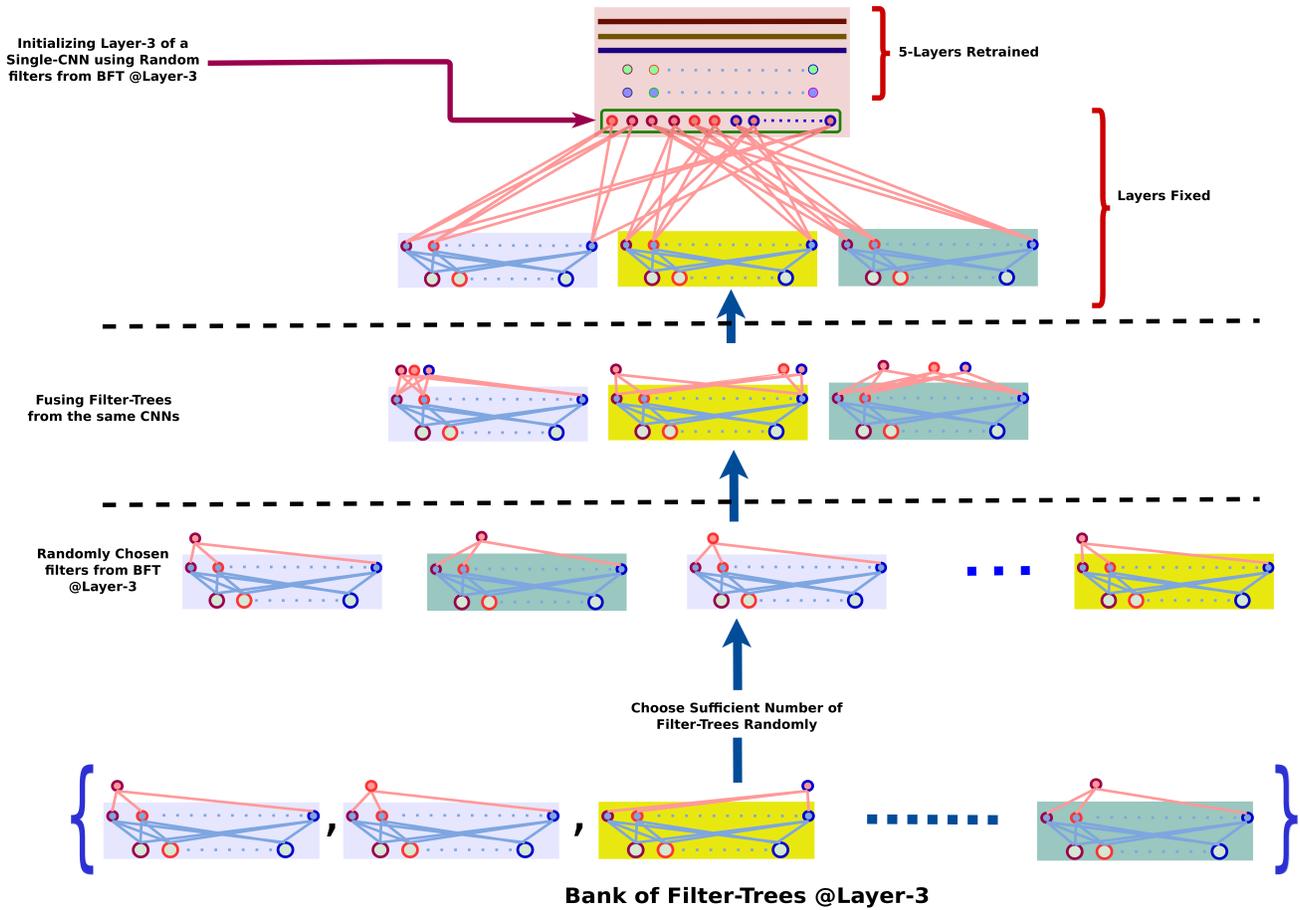}
  \center{\caption{Illustrating the initialization of the new target CNN from the BFT using layer-3 filters. First choose
  sufficient number of filter-trees (from the BFT) as needed at the layer-3 of target net. If multiple filter-trees come from the same source 
  CNN, fuse the \textit{layers 1 and 2} of the filter-trees
  coming from the same CNNs. (Fusing conserves the memory and reduce the forward-pass computation). After this add new layers on top of layer-3 as needed for the target CNN 
   architecture and train only the newly added layers keeping 
  \textit{3-convolutional} layers fixed. \textit{(Note:This figure is best viewed in color)} \label{fig:fig_BFT_init}}}
\end{center}
\end{figure*}

This construction of the target CNN using a BFT is schematically illustrated in Figure~\ref{fig:fig_BFT_init}.

\section{Experimental Setup}
\label{sec:expt-setup}
We conduct a number of simulation experiments to show the effectiveness of BFT for learning CNNs. For all our experiments the CNN architecture we consider is 
the AlexNet \cite{Krizhevsky} implemented in caffe framework~\cite{caffe}. AlexNet comprises of five convolutional 
layers and 3 fully-connected (FC) layers. 

The datasets we consider are all from Imagenet~\cite{imagenet}. 
All our experiments are conducted using five non overlapping subsets of ImageNet. 
 Each subset consists of 10 classes each spread across terrestrial, aquatic, indoor and random object categories.
The subsets are referred to as set-1 to set-5 where set-1 has classes 1 to 10, set-2 has classes 491 to 500, set-3 has 
classes 501 to 510, set-4 has classes 991 to 1000 and set-5 comprises random ten classes excluding the ones included in set-1 to set-4. 

CNNs trained using the normal protocol for training Alexnet are used as the source networks for creating the BFT. 
In each experiment, the CNNs trained on four subsets are used for building the BFT. A target net constructed from this BFT is then used for learning the remaining subset. In the initial set of experiments, we construct the BFT using layer-3 filters in the learnt CNNs. In the target net, the CNN layers subsequent to the BFT layer (that is, layer-3) are trained using the training examples from the 
target  data set (while the weights in the first three layers are fixed by sampling from the BFT). This CNN is compared against the CNN which is trained from scratch (that is, using random initialization of weights). We also compare the BFT-based CNN with a CNN learnt through conventional transfer learning. For this, we start with one of the four CNNs learnt earlier and re-learn the weights of this CNN only from layer-4 onwards using target data set. We keep the transferred weights (that is, weights in the first 3 layers) fixed like this in conventional (or network-level) transfer learning, so as to get a fair comparison with BFT-based transfer learning.     

The layer-3 of Alexnet has 384 filters. Since we construct BFT using four learnt CNNs, we have a total of 1536 filter-trees in the BFT. Out of this, we randomly select 384 filter-trees to construct the target net as explained earlier. Since we learn weights only in layers 4 to 8, the computation involved in any training epoch is less than that for normal training and is same as that in conventional transfer learning without fine tuning. Hence, to compare speed of learning, we compare the number of iterations needed to reach a level of accuracy. 
As mentioned earlier, we first consider BFT constructed using layer-3 filters in learnt CNNs. Later, we also present results with BFT constructed at the level of layer-4 or layer-5.

\begin{table*}[htp]
   \begin{center}
    
    \begin{tabular}{ | l | l | l | l | l | l | }
    \hline
    {\textbf{Type}} & \textbf{set-1} & \textbf{set-2}  & \textbf{set-3} & \textbf{set-4} & \textbf{set-5}  \\ \hline \hline
    \textbf{Normal-train} & \textbf{0.73} & \textbf{0.81} & \textbf{0.70} &  \textbf{0.74} & \textbf{0.85} \\ \hline \hline
    \textbf{conv3-BFT}&\textbf{0.73\footnotesize{$\pm$0.016}}&0.79\footnotesize{$\pm$0.01}&\textbf{0.7\footnotesize{$\pm$0.01}}&\textbf{0.75\footnotesize{$\pm$0.01}}	&\textbf{0.85\footnotesize{$\pm$0.08}}	\\ \hline \hline
    \textbf{conv3-Net}	&	0.70	\footnotesize{$\pm$	0.015}	&	0.76	\footnotesize{$\pm$	0.01}	&	0.67	\footnotesize{$\pm$	0.02}	&	0.73	\footnotesize{$\pm$	0.03}	&	0.82	\footnotesize{$\pm$	0.01}	\\ \hline
    \end{tabular}
    \caption{\label{tab:table-1} Comparing average accuracies of CNNs trained normally from scratch (normal-train) with conventional transfer learning (conv3-Net) and BFT-based transfer learning (conv3-BFT).
  Results are for the five subsets, set-1 to set-5 of ImageNet.}
\end{center}
\end{table*}

\begin{table*}[htp]
   \begin{center}
    
    \begin{tabular}{ | l | l | l | l | l | l | }
    \hline
    {\textbf{Type}} & \textbf{set-1} & \textbf{set-2}  & \textbf{set-3} & \textbf{set-4} & \textbf{set-5}  \\ \hline \hline
    \textbf{conv3-BFT}	&	\textbf{0.73	\footnotesize{$\pm$	0.016}}	&	\textbf{0.79	\footnotesize{$\pm$	0.01}}	&	\textbf{0.7	\footnotesize{$\pm$	0.01}}	&	\textbf{0.75	\footnotesize{$\pm$	0.01}}	&	\textbf{0.85	\footnotesize{$\pm$	0.08}}	\\ \hline
    \textbf{conv3-Net}	&	0.70	\footnotesize{$\pm$	0.015}	&	0.76	\footnotesize{$\pm$	0.01}	&	0.67	\footnotesize{$\pm$	0.02}	&	0.73	\footnotesize{$\pm$	0.03}	&	0.82	\footnotesize{$\pm$	0.01}	\\ \hline  \hline
    \textbf{conv4-BFT}	&	0.72	\footnotesize{$\pm$	0.016}	&	0.76	\footnotesize{$\pm$	0.02}	&	0.67	\footnotesize{$\pm$	0.01}	&	0.73	\footnotesize{$\pm$	0.01}	&	0.83	\footnotesize{$\pm$	0.01}	\\ \hline
    \textbf{conv4-Net}	&	0.69	\footnotesize{$\pm$	0.019}	&	0.73	\footnotesize{$\pm$	0.01}	&	0.62	\footnotesize{$\pm$	0.01}	&	0.71	\footnotesize{$\pm$	0.03}	&	0.80	\footnotesize{$\pm$	0.01}	\\ \hline  \hline
    \textbf{conv5-BFT}	&	0.69	\footnotesize{$\pm$	0.010}	&	0.73	\footnotesize{$\pm$	0.01}	&	0.64	\footnotesize{$\pm$	0.02}	&	0.69	\footnotesize{$\pm$	0.01}	&	0.80	\footnotesize{$\pm$	0.01}	\\ \hline 
    \textbf{conv5-Net}	&	0.68	\footnotesize{$\pm$	0.031}	&	0.72	\footnotesize{$\pm$	0.04}	&	0.61	\footnotesize{$\pm$	0.06}	&	0.67	\footnotesize{$\pm$	0.07}	&	0.78	\footnotesize{$\pm$	0.04}	\\ \hline

    \end{tabular}
    \caption{\label{tab:table-3} Comparison of BFTs constructed  at layers 3, 4 and 5.  The the BFT at the layer-level 3 has the best performance.}
\end{center}
\end{table*}

\begin{table*}[htp]
   \begin{center}

    \begin{tabular}{ | l | l | l | l | l | l | }
    \hline
    {\textbf{Type}} & \textbf{set-1} & \textbf{set-2}  & \textbf{set-3} & \textbf{set-4} & \textbf{set-5}  \\ \hline \hline
    \textbf{Network transfer (w.o shuffle)}	&	0.68	\footnotesize{$\pm$	0.031}	&	0.72	\footnotesize{$\pm$	0.04}	&	0.61	\footnotesize{$\pm$	0.06}	&	0.67	\footnotesize{$\pm$	0.07}	&	0.78	\footnotesize{$\pm$	0.04}	\\ \hline
    \textbf{Network transfer (w. shuffle; src and tgt same)}	&	0.65&	0.60&	0.45&	0.56&	0.68 \\  \hline
    \textbf{Network transfer (w. shuffle; src and tgt diff.)}& 0.60\footnotesize{$\pm$	0.07}&0.48\footnotesize{$\pm$	0.08}&0.41\footnotesize{$\pm$	0.08}&0.52\footnotesize{$\pm$	0.01}&0.57	\footnotesize{$\pm$	0.07}	\\  \hline \hline
    \textbf{BFT}	&	0.69	\footnotesize{$\pm$	0.010}	&	0.73	\footnotesize{$\pm$	0.01}	&	0.64	\footnotesize{$\pm$	0.02}	&	0.69	\footnotesize{$\pm$	0.01}	&	0.80	\footnotesize{$\pm$	0.01}	\\ \hline 
    \end{tabular}
    \caption{\label{tab:table-2} Results of the experiments justifying the filter-tree as an unit of transfer instead of an individual filter.}
\end{center}
\end{table*}

\section{Performance of BFT-based CNNs}
\label{sec:results}

 Table-\ref{tab:table-1} compares accuracies obtained with the normally trained CNN against BFT-based CNN and with conventional transfer learning. Normally trained CNN refers to the case where we start with random initialization and learn all weights using the training data. For BFT-based CNN (denoted as `conv-3 BFT', in the table), the net used to learn any one of the data sets is based on BFT constructed using CNNs trained on the other sets as explained earlier. The BFT is constructed using layer-3 filters. 
For CNN based on conventional transfer learning (denoted as 'conv3-Net', in the table), we start with a CNN learnt on one of the other data sets. As explained earlier, for both BFT-based and conventional transfer learning, the weights in the first three layers are fixed and we learn weights from layer-4 onwards using the training data of the new task.  
The table shows the average accuracy and the standard deviation over five random trials of the two transfer learning methods.

The results clearly show the performance superiority of the BFT-based method over conventional transfer learning. The accuracy achieved by conventional transfer learning (with transferred weights fixed) is always significantly less than that of normally trained CNN. The BFT-based CNN, where the transferred weights are  fixed, achieves the same performance as normally trained CNNs. This may be because, in the BFT-based CNNs, the transferred filters can come from different source networks thus giving it sufficient diversity of features for the new task. This experiment establishes clearly that BFT-based transfer learning is better than conventional transfer learning in CNNs.

\begin{figure}
\begin{center}
  \includegraphics[width=0.308\textwidth]{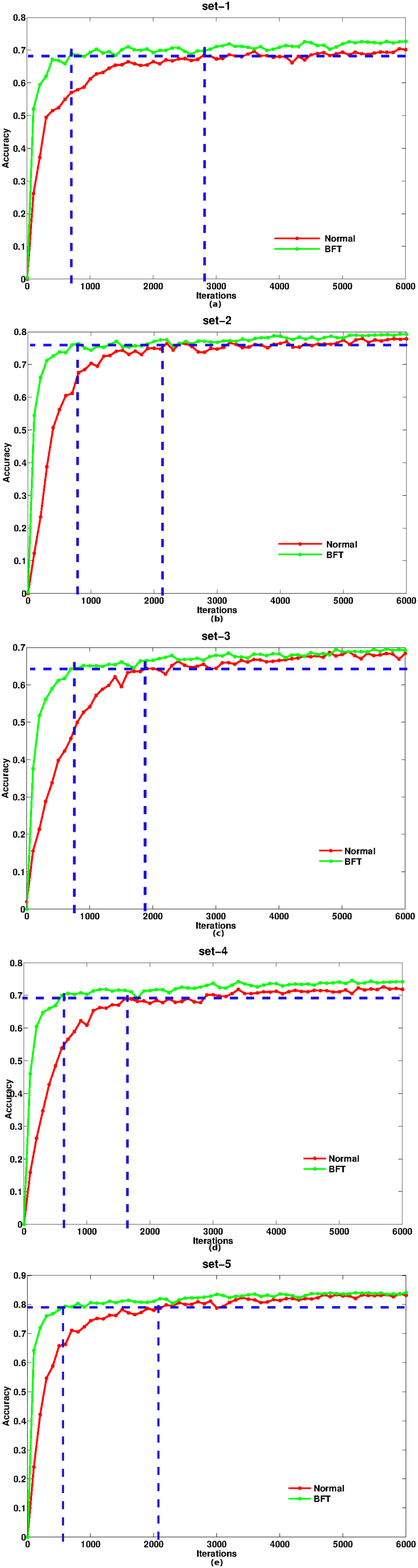}
  \center{\caption{Illustrating iterations taken for convergence by the normally trained CNNs and BFT based CNNs for subsets \textit{set-1 to set-5}. 
  The BFT initialized CNNs learn at least twice as fast.
  \textit{(Note:This figure is best viewed in color)} \label{fig:fig_BFT_conv}}}
\end{center}
\end{figure}
As mentioned earlier, the main motivation for transfer learning is to improve efficiency of learning CNNs for new tasks. To this end, we compare the convergence rate of BFT-based transfer learning against that of learning the CNN normally in Figure~\ref{fig:fig_BFT_conv}. In the figure we mark the iterations needed in the two cases to reach the same level of accuracy. As can be seen from the figure, the BFT-based method takes less than half the iterations needed by the normal way of training CNNs starting with random weights. Thus, both in terms of speed and accuracy, the BFT proposed here is a better learning mechanism.

\subsection{BFT at different layers}
All the experiments described so far used BFT constructed from layer-3 filters. Here we explore BFT using layer-4 and layer-5 filters. As explained earlier, if we use layer-$k$ filters for BFT, then, in the target net all weights up to layer-$k$ are fixed and only weights from layer-$(k+1)$ onwards are learnt using new training data. We compare these different BFT-based transfer learning of CNNs with conventional transfer learning with the same constraints on learning of weights. Thus, for conventional transfer learning at layer-$k$, we start with one of the learnt CNNs, and, keeping all weights up to layer $k$ fixed,  learn weights only from layer $(k+1)$ onwards (where is $k$ is chosen to be either 3 or 4 or 5).

The results of the experiments are provided in the Table-\ref{tab:table-3}. (In the table `conv$4$-BFT' refers to BFT-based transfer learning with layer-$4$ filters and `conv$4$-net' refers to conventional transfer learning with weights up to layer-$4$ fixed; and similarly for other layers). The results in the table show that BFT-based transfer learning is better than conventional transfer learning  at all the levels of layers 3, 4 and 5. This amply demonstrates the effectiveness of the idea of BFT proposed here.  The results also show that BFT-based transfer learning gives best results when the BFT is constructed using layer-3 filters. This is understandable because it is only such intermediate level of features that are likely to be very useful across different tasks.

\subsection{Relevance of filter-tree as the unit for transfer learning}
The idea of BFT proposed in this paper has two issues that are important. One is that we sample from multiple source networks for the transfer and this gives enough diversity in the transferred features which contributes to better performance. The second and more important issue is that the weights are transferred from source networks to the target network in terms of filter-trees. This, as we showed here, is very effective; with the transferred weights kept fixed, we get same performance as that of CNN learnt from scratch. The idea of making a bank of filters from many source networks and then sampling from it to train a new target network has been proposed earlier in~\cite{kirthi}. There the  units of transfer are individual filters or vectors of weights connecting to a node in any layer. In this method the target network, constructed from a bank of filters, needs finetuning (which is same as relearning) of transferred weights for good performance. Without such re-learning of transferred weights (along with other weights in the network) the accuracy drops. Thus, our simulation results here show that filter-tree provides the correct way to transfer learnt features to a new task.

Consider a node (filter) in the layer-$k$ of a CNN. While it represents a learnt feature, the weights from layer-$(k-1)$ to this node do not completely specify the filter; which weight connects to which filter in the previous layer is important for this feature detector to perform well. Thus, if we keep the weight vector of this node fixed but shuffle the filters in the previous layer, then this node can no longer represent the learnt feature. On the other hand, the filter-tree that we defined here contains the full context and hence represents the learnt feature in a proper manner. This is the reason why the filter-tree is much more effective as a unit of transfer. 

In Table~\ref{tab:table-2} we provide some empirical results to support the above argument. This table contains results of experiments with conventional transfer learning and BFT at the level of layer-5.  The first row of the table, termed network transfer (w.o. shuffle), refers to conventional transfer learning that we have been considering. Here, for each set, we start with the CNN learnt with one of the other sets and then re-learn the weights beyond layer-5. In the next two rows of the table we consider the conventional transfer learning with shuffle. That is, here we start with a pre-learnt network, randomly shuffle filters in layers below layer-5 and then learn weights only above layer-5. In the second row of the table, we show the case where source and target tasks are same. That is, we learn a CNN for, say, set-1, shuffle filters below layer-5, and then retrain weights above layer-5 on the same task, namely, set-1. Without shuffle, such a transfer learning would be perfect. However, as can be seen from the table, because of the shuffle, the performance drops drastically. This, for example. shows why  choosing from a bank of weight vectors corresponding to individual filters may not be really effective. The third row of table~\ref{tab:table-2} shows the case when we do conventional transfer learning with shuffle and where the source and target tasks are different. As one can expect, the performance here is worse than that in the second row of the table. The last row shows the performance obtained with BFT-based transfer learning. The results presented in table~\ref{tab:table-2} provide a good justification for filter-tree being a good choice for subnetwork level of transfer.

\section{Conclusion}
\label{sec:conc}
While CNNs are very useful in many applications, learning of CNNs needs large number of training examples and is computationally intensive. Transfer learning has been explored as a way of improving the efficiency of learning CNNs. In conventional transfer learning one essentially starts with a previously learnt network and finetunes the weights using data for new task. Here transfer of learnt knowledge is viewed as transfer of individual weight vectors. Such transfer learning does not achieve good accuracy unless we finetune or re-learn all the weights. 
In this paper we proposed a new mechanism of transfer learning using what we called bank of filter-trees~(BFT). 

CNNs are believed to be very effective in many pattern recognition tasks because they automatically learn relevant features from the training data. Thus transferring learnt knowledge should mean reusing previously learnt features and developing new features for the target task using previously learnt features. Thus transfer of learnt knowledge should be at the level of feature generators (or subnetworks) rather than at the level of individual weight vectors. This is the motivation behind the proposed filter-trees which are subnetworks at correct level of granularity.

The BFT is an ensemble of such filter trees from many pre-learnt CNNs. Thus BFT allows us to effect transfer from any number of source networks and it also naturally allows for a rich set of features to sample from while constructing the target network. 

Through many simulation experiments, we showed that BFT-based transfer learning achieves same accuracy as normal learning of CNNs while needing less than half the time. This is achieved with no finetuning of the transferred weights. We also showed that conventional transfer learning cannot achieve similar accuracies.

An attractive feature of the idea of BFT is that, given a rich enough set of filter-trees, we can learn new CNNs using these filters as the initial features. In the target CNN, all the weights in the filter-tree are fixed. This means that at the level of feature learning we can effect learning of new tasks without forgetting the old ones. This is an important direction in which the idea of BFT can be further explored. 

In this paper, while constructing the target net using the BFT we have simply uniformly sampled from BFT. However, it is more attractive to have a method of sampling from BFT so as to minimize redundancy and maximize diversity in the set of chosen filter-trees. This is another important direction in which the work presented here can be extended.

\newpage
\bibliography{kirthi_arxiv_bib}
\bibliographystyle{aaai}

\end{document}